\documentclass[runningheads]{llncs}
\usepackage{graphicx}
\usepackage{bbding}
\usepackage{float}
\usepackage[export]{adjustbox} 
\usepackage{ragged2e} 
\usepackage{booktabs,makecell, multirow, tabularx}

\begin{document}
\title{MuraNet: Multi-task Floor Plan Recognition with Relation Attention}
%
%

\author{Lingxiao Huang\inst{(}\Envelope\inst{)}
\and Jung-Hsuan Wu
\and Chiching Wei
\and Wilson Li
}

%
%
\institute{ Foxit Software, Fremont, CA 94538, USA 
\email{\{lingxiao\_huang, matt\_wu, jeremy\_wei, wensheng\_li\}@foxitsoftware.com}\\}
%

\maketitle              
\begin{abstract}
The recognition of information in floor plan data requires the use of detection and segmentation models. However, relying on several single-task models can result in ineffective utilization of relevant information when there are multiple tasks present simultaneously. To address this challenge, we introduce MuraNet, an attention-based multi-task model for segmentation and detection tasks in floor plan data. In MuraNet, we adopt a unified encoder called MURA as the backbone with two separated branches: an enhanced segmentation decoder branch and a decoupled detection head branch based on YOLOX, for segmentation and detection tasks respectively. The architecture of MuraNet is designed to leverage the fact that walls, doors, and windows usually constitute the primary structure of a floor plan's architecture. By jointly training the model on both detection and segmentation tasks, we believe MuraNet can effectively extract and utilize relevant features for both tasks. Our experiments on the CubiCasa5k public dataset show that MuraNet improves convergence speed during training compared to single-task models like U-Net and YOLOv3. Moreover, we observe improvements in the average AP and IoU in detection and segmentation tasks, respectively.
Our ablation experiments demonstrate that the attention-based unified backbone of MuraNet achieves better feature extraction in floor plan recognition tasks, and the use of decoupled multi-head branches for different tasks further improves model performance. We believe that our proposed MuraNet model can address the disadvantages of single-task models and improve the accuracy and efficiency of floor plan data recognition.

\end{abstract}
\section{Introduction}
Architectural floor plan data are standardized data that are used to make the design, construction, and other related work convenient. There are stringent requirements for the recognition accuracy of objects with different design components when automatically identifying design drawing data, to reduce errors between different works around the same floor plan data.
There are several factors that makes the floor plan recognition difficult such as:
(1) Special data characteristics: These data are composed of simple lines; thus, few features exist in aspects such as texture and color, and most features focus on the shape, structure, and relationships among different parts.
(2) Complex semantic relationships exist between different objects.
(3) Approaches such as detection and segmentation tasks usually work best for subsets of objects differently.

By looking into the literatures, classical convolutional neural network (CNN) models have primarily been used for different visual tasks on floor plan datasets. For example,  the fully convolutional network (FCN), U-Net~\cite{ref_lncs12}, HRNet~\cite{ref_lncs27}, DeepLab~\cite{ref_lncs28}, and Mask R-CNN~\cite{ref_lncs29} have been applied for mainframe and layout parsing, whereas YOLO and Faster R-CNN~\cite{ref_lncs30} have been used for component detection. However, these models do not consider the characteristics of floor plan data, including few low-level features and strong correlations among different objects at a high level. In recent years, attention mechanisms have been applied to vision tasks. Several state-of-the-art models have achieved success on many well-known public vision datasets, such as DETR~\cite{ref_lncs25}, ViT~\cite{ref_lncs9}, and the Swin Transformer~\cite{ref_lncs26}. Attention mechanisms can focus on the relationships among the data and improve model accuracy, which is beneficial for the complex relationships between different object types and recognition tasks in floor plan data. Because of the high-level features of floor plan data, a strong correlation exists between the different components, which makes it possible to use attention mechanisms in recognition tasks on floor plan data.

\begin{figure}
\includegraphics[width=\textwidth]{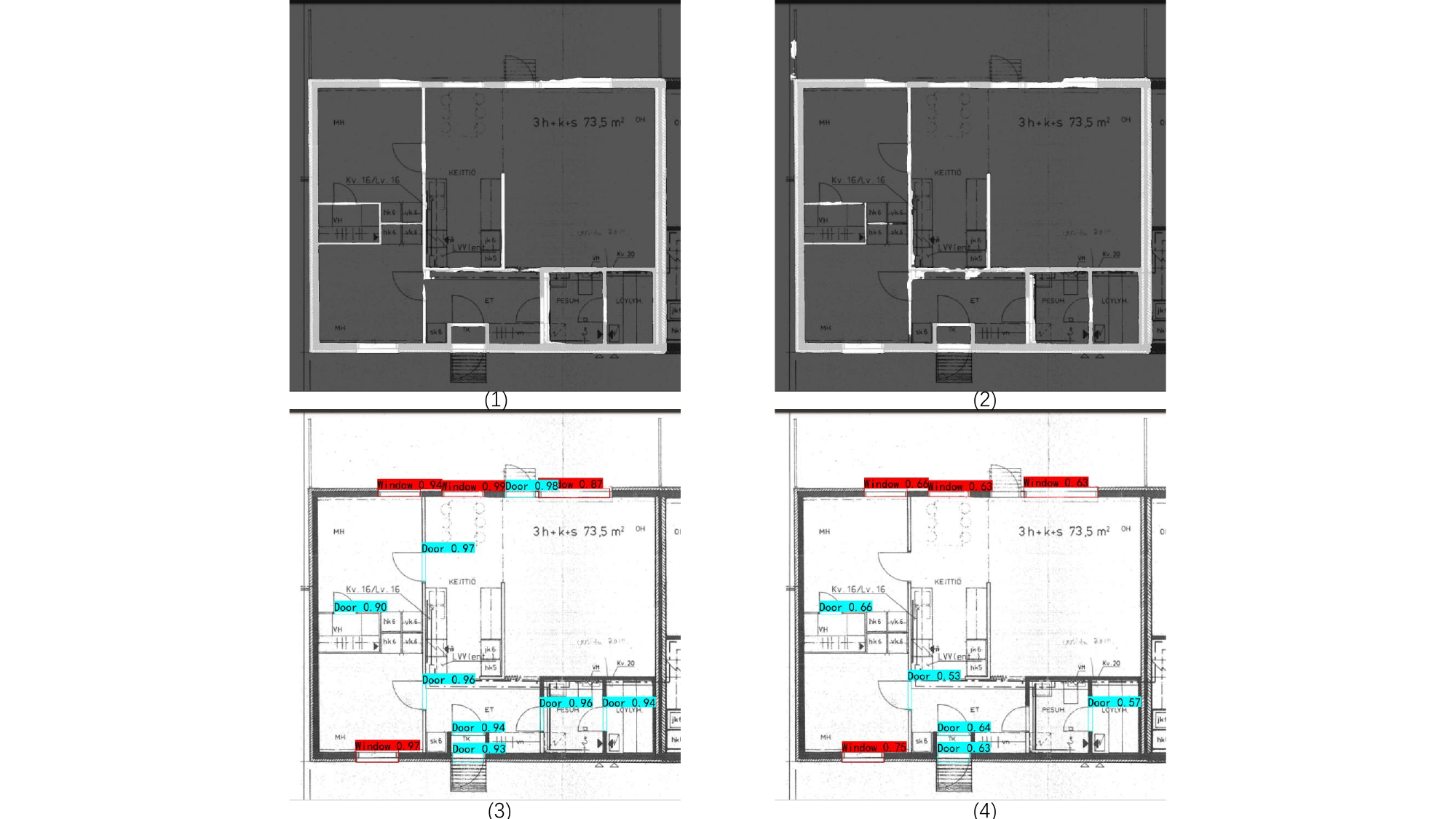}
\caption{Visualization results of a. MuraNet and b. U-Net + YOLOv3. (1) visualizes the MuraNet segmentation results, (2) visualizes the segmentation results with U-Net(D5), (3) visualizes the MuraNet detection results, (4) visualizes the detection results with YOLOv3. We can see that the segmentation results of MuraNet have fewer misidentifications and are more accurate than those of U-Net(D5). MuraNet is able to accurately and completely detect each door and window, but YOLOv3 misses two doors.} \label{fig11}
\end{figure}

In this study, we propose MuraNet, which is an attention-based multi-task model for segmentation and detection tasks in floor plan analysis. We adopt a unified encoder with the MURA module as the backbone. The improved segmentation decoder branch and decoupled detection head branch from YOLOX~\cite{ref_lncs15} are used for the segmentation and detection tasks, respectively. 

Our contributions are twofold. 
First, our proposed MuraNet jointly considers the wall pixel-level segmentation and doors and windows vector-level detection at the same time.
Second, we add attention mechanism for the model to leverage the correlations between walls, doors and windows to enhance the accuracy. Because walls, doors, and windows usually jointly constitute the main frame of a floor plan's architecture, we believe the joint training will provide advantages.
The performance of our model is excellent in terms of experimental values and intuitive visualization, as illustrated in Fig.~\ref{fig11}.

This article is structured as follows:
First, we introduce classical models for the recognition task and their applications to floor plans in the related work section. We also analyze the feasibility of applying the attention mechanism and explain its function and principle in this context.
Next, in the methods section, we provide a detailed description of the MuraNet model architecture, including the attention module, the unified backbone, and multi-head branches.
Then, in the experimental section, we present comparative experiments with classical models and conduct ablation experiments for the attention module and decoupled branch.
Finally, we conclude by summarizing the contributions of this work and proposing future research directions while also acknowledging the current limitations.

\section{Related Works}

\subsection{Recognition Tasks}
In recognition tasks, traditional CNN models are widely used and stable. Different models are often applied to different recognition tasks in floor plans: scene segmentation models are applied to mainframe and layout parsing, object detection models are applied to component detection and recognition, and other models are used for specific tasks. For example, the graph convolutional network model~\cite{ref_lncs24} has been used for the analysis of vector data such as CAD and PDF, whereas generative adversarial network models have been used to analyze the main structure of drawings in a data-generated manner.

\subsubsection{Detection Models.} 
Many graphic symbols, such as doors, sliding doors, kitchen stoves, bathtubs, sinks, and toilets, need to be recognized in floor plans. Faster R-CNN includes an end-to-end model, and region proposal-based CNN architectures for object detection are often used to identify the parts and components in floor plans~\cite{ref_lncs2,ref_lncs4}. The YOLO model, which is a fast one-stage convolutional method that automatically detects structural components from scanned CAD drawings, was proposed in~\cite{ref_lncs22}. Several improved versions of the YOLO model, such as YOLOv2, have also been applied extensively~\cite{ref_lncs23}.

\subsubsection{Segmentation Models.} 
FCNs have exhibited effective performance in segmentation tasks in the early stages~\cite{ref_lncs2}. U-Net~\cite{ref_lncs12} and the variants of U-Net have been designed based on this concept and have been proven to be outstanding in image segmentation tasks. Mask R-CNN is an instance segmentation model that is used to detect building and spatial elements~\cite{ref_lncs5}.

In some works, they use separate detection and segmentation models to identify different objects. In some multi-task works~\cite{ref_lncs6}, they use same backbone such as VGG-16 for feature extraction, a decoder such as U-Net~\cite{ref_lncs12} for segmentation, and a detector such as SSD for detection to achieve detection and segmentation tasks simultaneously. But recognition tasks are not independent of one another, and a relationship exists among recognition targets because they constitute a complete floor plan. In some multi-task works~\cite{ref_lncs8,ref_lncs19}, the outputs of these models mentioned are pixel-level segmentation maps, but the outputs of our model are pixel-level segmentation maps and vector-level detection coordinates. None of these studies considered the interrelationships among different recognition targets.
\subsection{Attention Networks} Attention mechanisms enable a network to focus on important parts adaptively. Recent models with attention mechanisms have exhibited high performance in image classification and dense prediction tasks such as object detection and semantic segmentation---for example, DETR~\cite{ref_lncs25}, ViT~\cite{ref_lncs9}, and the Swin Transformer~\cite{ref_lncs26} based on self-attention, as well as HRNet+OCR, VAN~\cite{ref_lncs10}, and SegNeXt~\cite{ref_lncs11} based on convolution attention. Several state-of-the-art models have used a similar approach on many well-known public computer vision datasets. 

\subsubsection{Self-Attention.} Self-attention mechanisms are primarily concerned with the overall relationships of all data. The self-attention mechanism first appeared in the transformer model in the natural language processing (NLP) field and is currently an indispensable component of NLP models. Several self-attention-based methods perform similarly to or even better than CNN-based methods. These models break down the boundaries between the computer vision and NLP fields, thereby enabling the development of unified theoretical models.

\subsubsection{Convolution Attention.} Convolution attention mechanisms are primarily concerned with the relationships among data that are processed by different convolution kernels. Convolution attention includes two categories: spatial attention and channel attention. Spatial attention focuses on the mutual relationships based on spatial information, whereas channel attention aims to select important objects for the network adaptively. In CNN models, a large kernel convolution layer is used to build both spatial and channel attention, which is known as a large kernel attention (LKA) mechanism. Both VAN~\cite{ref_lncs10} and SegNeXt~\cite{ref_lncs11} use the LKA mechanism, with SegNeXt being the most relevant reference for our model.

\section{Our Method}

\subsection{Model Architecture}

\begin{figure}
\begin{center}
\includegraphics[width=0.6\textwidth]{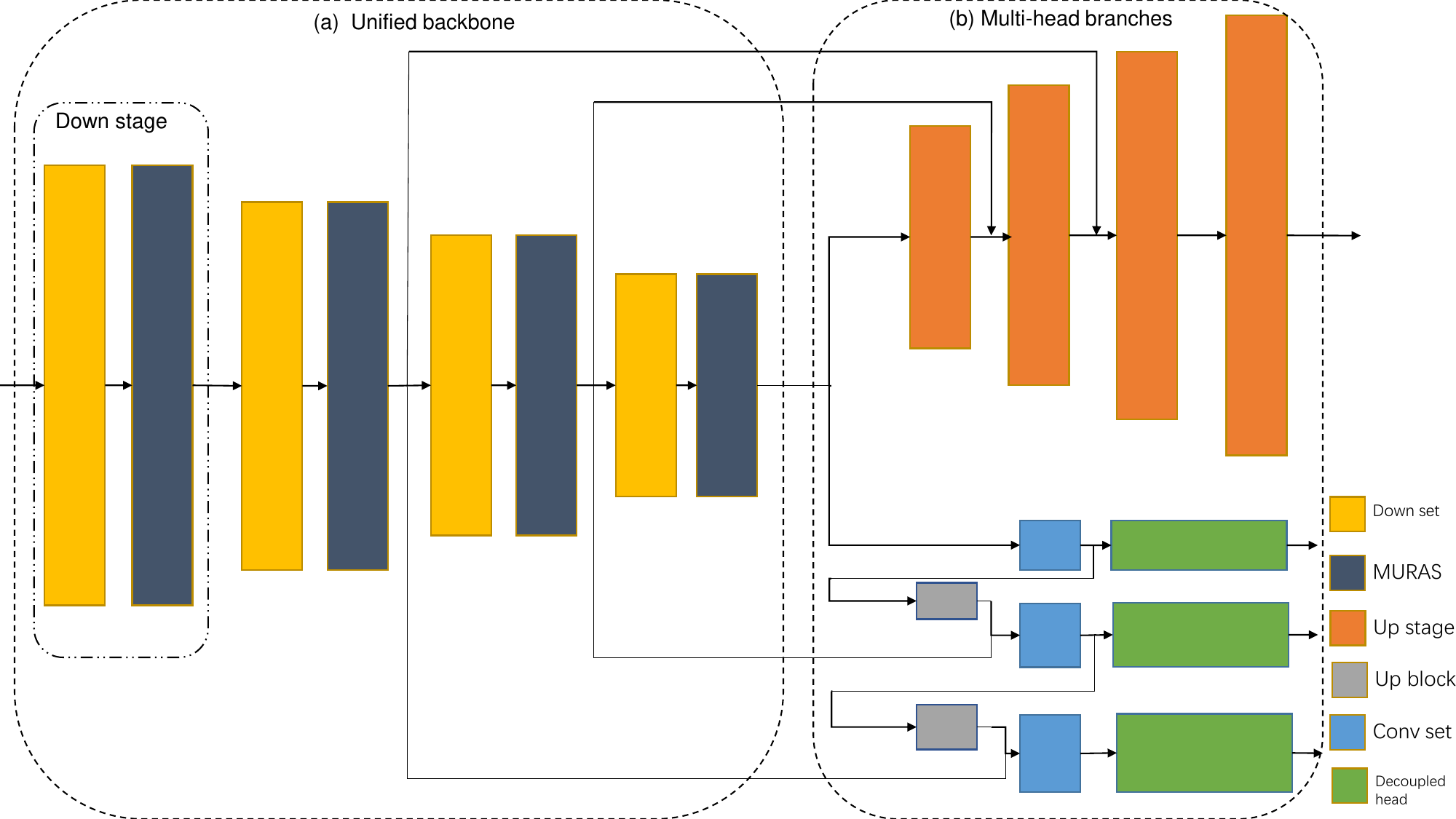}
\end{center}
\caption{MuraNet architecture uses (a) a unified backbone with the attention module as the encoder and (b) multi-head branches for multiple tasks.} \label{fig7}
\end{figure}

In this section, we describe the architecture of the proposed MuraNet in detail. 
Our model consists of a backbone with attention module, segmentation decoder and decoupled detection head. The backbone consists of 4 down-stages, and each down-stage consists of a set of downsampling convolutions and MURA in attention module. The segmentation decoder consists of 4 upsample-convolutional layers, and the features of last three stages will be kept and processed by the topmost convolutional layer after being aggregated. The decoupled detection head divides branches into classification and regression, and the regression branch divides branches into coordinate and IoU.
We adopt a unified architecture, as illustrated in Fig.~\ref{fig7}, which incorporates an attention-based relation attention module and uses multi-head branches for multiple tasks.

\subsubsection{Relation Attention Module.} 

\begin{figure}
\begin{center}
\includegraphics[width=0.6\textwidth]{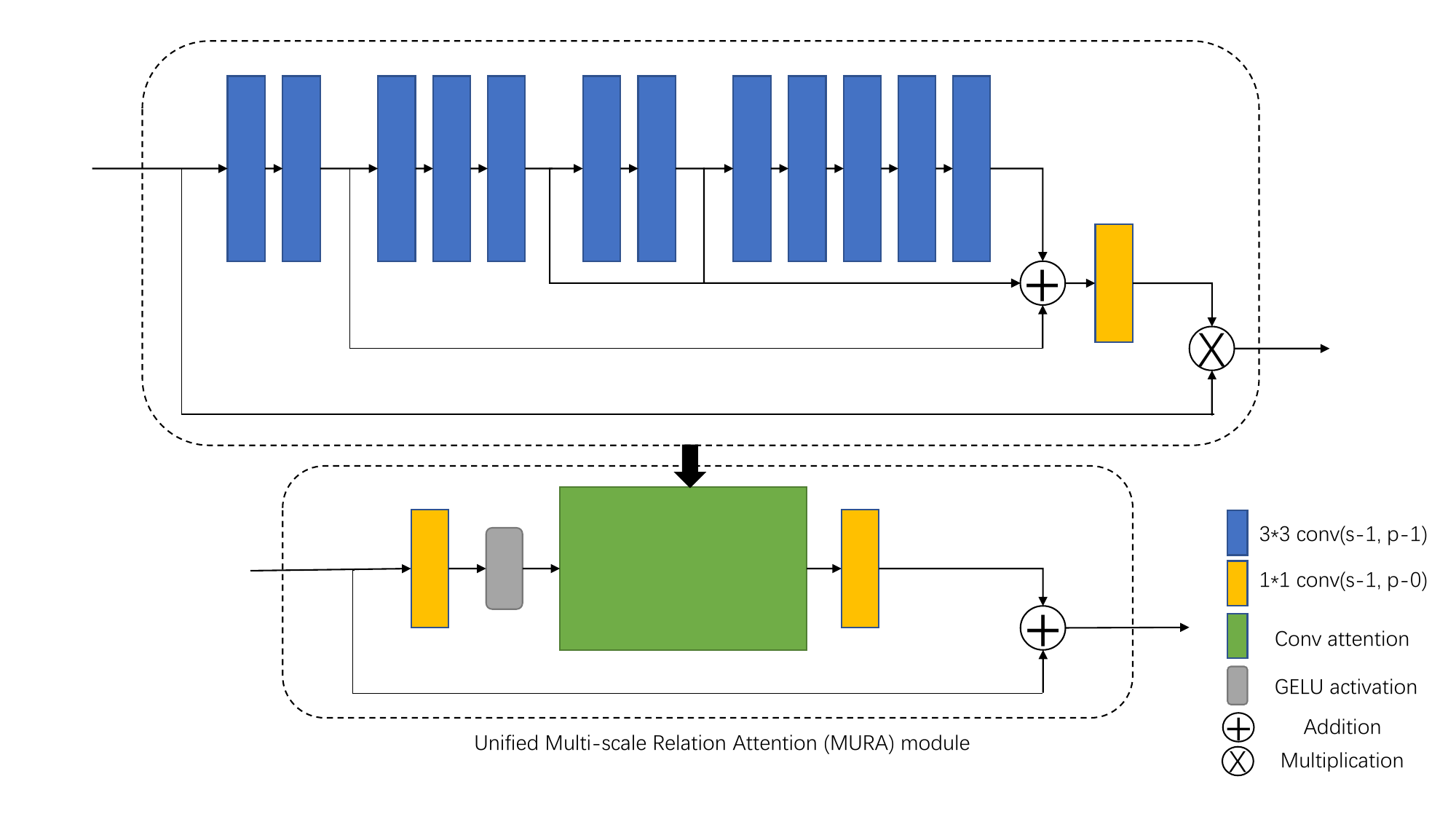}
\end{center}
\caption{MURA architecture uses (a) a series of $3\times 3$ convolution kernels instead of large-kernel depth-wise strip convolutions and (b) unified skip add connections to aggregate multi-scale features.} \label{fig3}
\end{figure}

We adopt a similar structure to that of ViT~\cite{ref_lncs9} and SegNeXt~\cite{ref_lncs11} to build our unified backbone as the encoder, but instead of multi-branch depth-wise strip convolutions, a unified multi-scale attention module is designed to focus on the relationships among multi-scale features. This module is known as the multi-scale relation attention (MURA) module. As depicted in Fig.~\ref{fig3}, the MURA module uses a series of $3\times 3$ convolution kernels instead of large-kernel depth-wise strip convolutions to avoid the sensitivity of specific-shaped convolution kernels to the detection target shape. Although most walls, doors, and windows are elongated in a floor plan, multi-angle directions exist, and most frames and components, such as house types, room shapes, and furniture, are square. The MURA module also uses skip add connections to aggregate multi-scale features~\cite{ref_lncs12}. Thus, the parameters can be shared and the associations among multi-scale features can be strengthened.

\subsubsection{Unified Backbone as the Encoder.} 
\begin{figure}
\begin{center}
\includegraphics[width=0.6\textwidth]{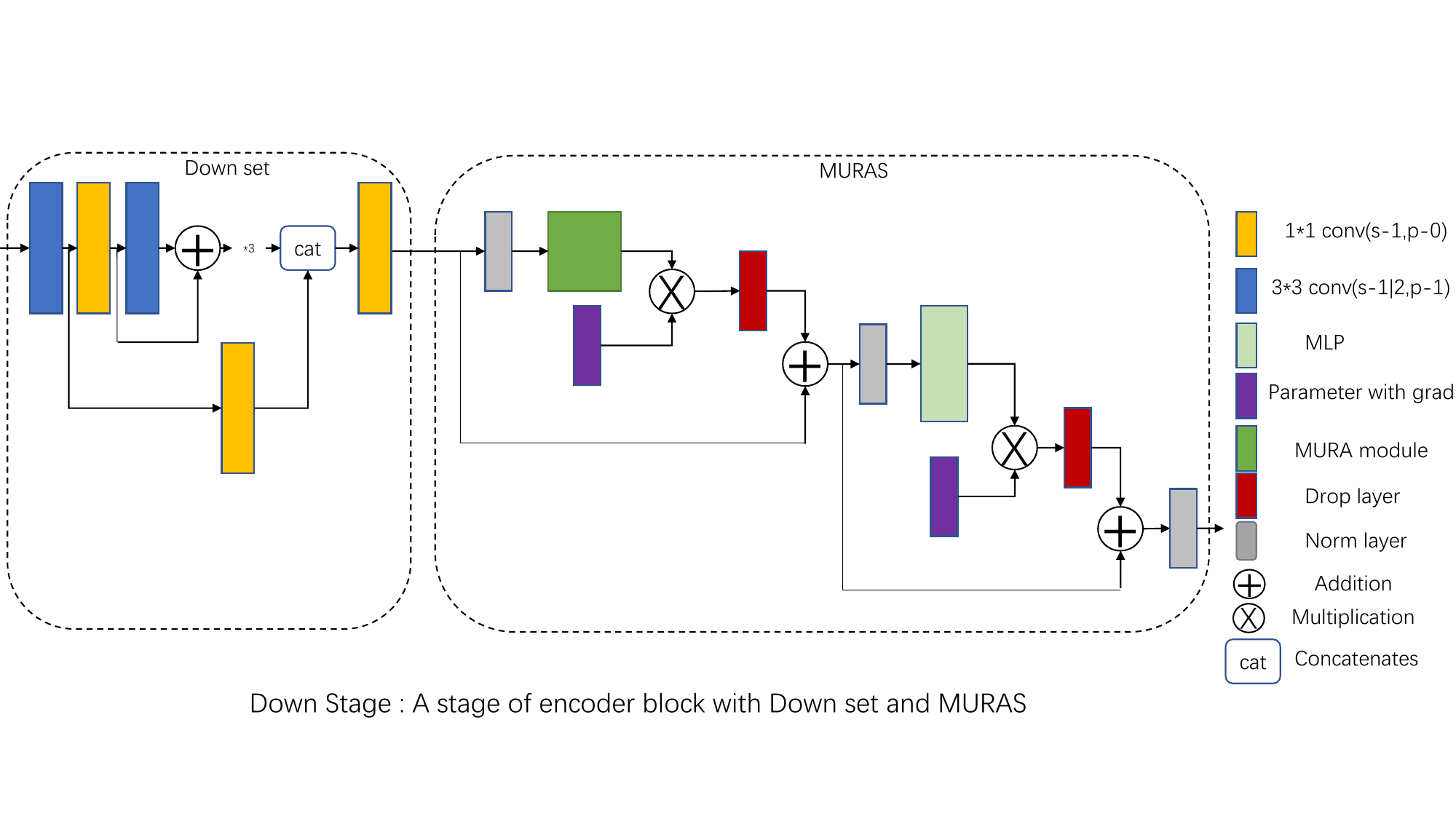}
\end{center}
\caption{Down Stage : A stage of encoder block with Down set and MURAS: two residual connections are used in one stage of the encoder block.} \label{fig2}
\end{figure}

Most components of floor plan data, such as walls, doors, and windows, are composed of simple lines, which means that few features exist in aspects such as texture and color. Furthermore, most features focus on the shape, structure, and relationships among different parts. As low-level feature information is insufficient, multi-level structured models and multi-scale feature aggregation are required to extract high-level features at different scales. We adopt the multi-level pyramid backbone structure for our encoder, following most previous studies~\cite{ref_lncs13,ref_lncs14}. 
Fig.~\ref{fig2} depicts a stage of the encoder block with the MURA module. We use two residual connections in one stage of the encoder block to avoid the vanishing gradient problem as the model depth increases.

\begin{figure}
\begin{center}
\includegraphics[width=0.6\textwidth]{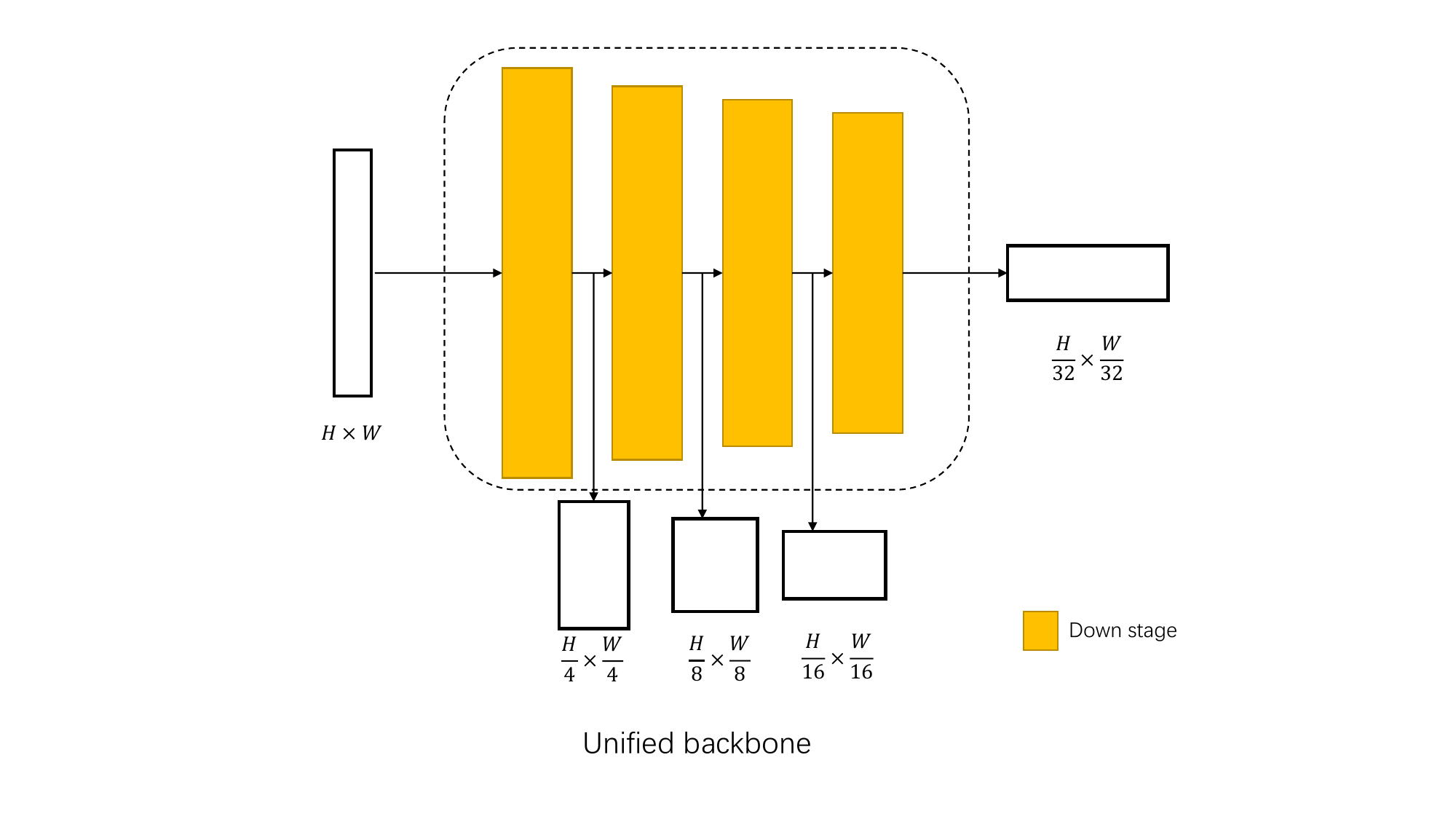}
\end{center}
\caption{Unified encoder architecture} \label{fig4}
\end{figure}

We adopt a robust four-level pyramid hierarchy that contains a downsampling block in each stage for the overall model architecture. The downsampling block uses a two-stride $3\times 3$ kernel convolution to decrease the spatial resolution, which is followed by a batch normalization layer. The model architecture is illustrated in Fig.~\ref{fig4}. The four resolutions are $\frac{H\times W}{4\times 4}$, $\frac{H\times W}{8\times 8}$, $\frac{H\times W}{16\times 16}$, and $\frac{H\times W}{32\times 32}$, where $ H\times W $ is the shape of the input data.

\subsubsection{Multi-Head Branches.} The input data are processed by a unified multi-level pyramid encoder to generate features of four different scales. The features of these different stages must be processed by heads that correspond to different tasks to obtain required data. We believe that it is necessary to use different head branches to process the corresponding tasks immediately following feature extraction through the unified backbone, because of the significant differences in the segmentation and detection tasks. We adopt multi-head branches for different recognition tasks, a decoder for the segmentation task, and a detector for the detection task.



\begin{figure}
\begin{center}
\includegraphics[width=0.6\textwidth]{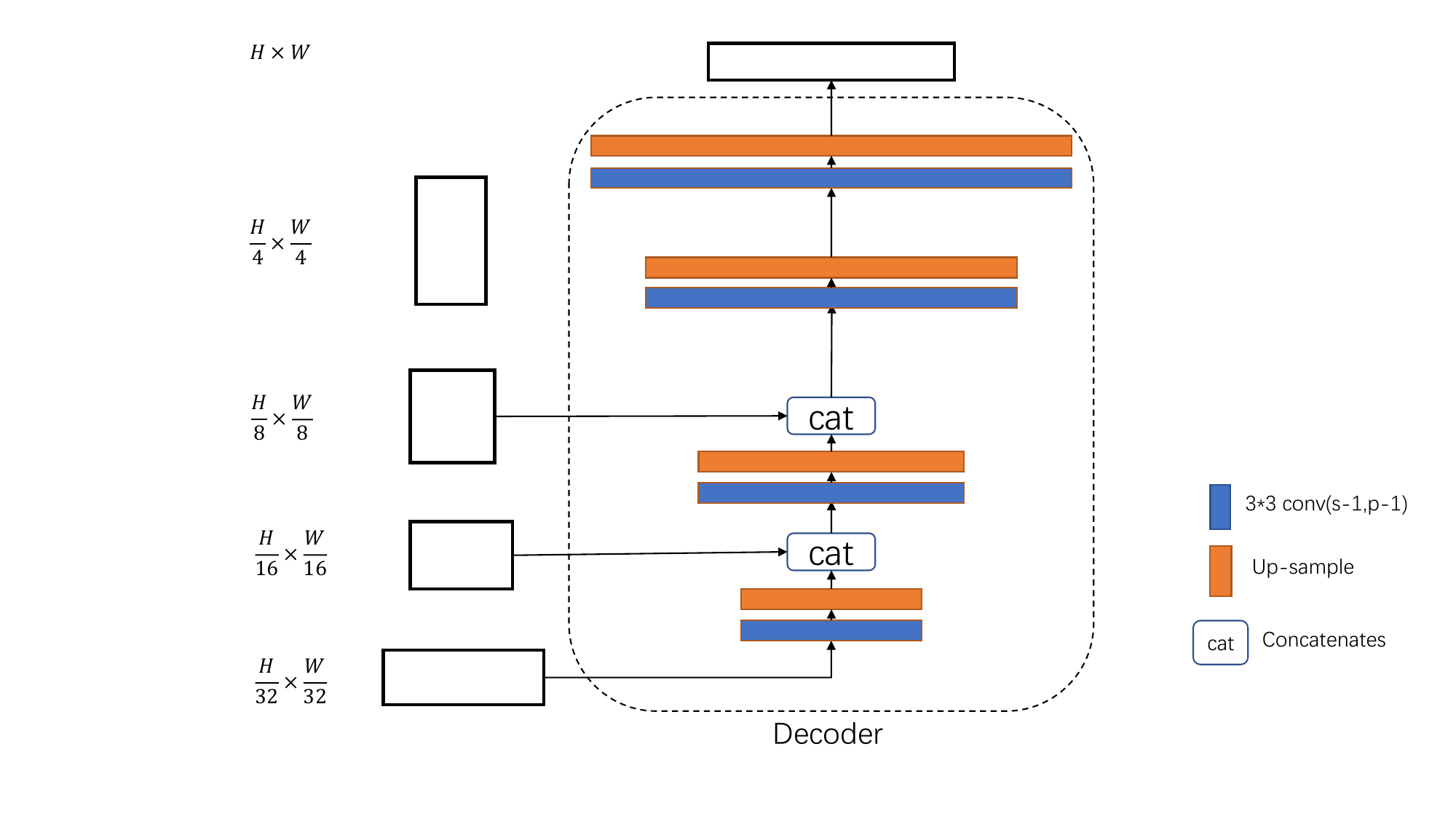}
\end{center}
\caption{Segmentation decoder} \label{fig5}
\end{figure}

\paragraph{Segmentation Decoder.} It is necessary for the decoder to aggregate and process multi-scale features in segmentation tasks. Thus, we adopted a powerful decoder based on SegNeXt~\cite{ref_lncs11} and U-Net~\cite{ref_lncs12}. 
SegNeXt uses a lightweight Hamburger, which  has been proven as an effective decoder, that can process features aggregated from the last three stages. Only the features of the last three stages are aggregated because the features from stage 1 contain excessive low-level information, which degrades the performance. The characteristics of the Hamburger decoder are consistent with those of the floor plan data; that is, few low-level features exist. 
In U-Net, the encoder--decoder structure is a symmetrical U-shape, and the model structure and skip connections that can connect multi-scale features have been demonstrated to be effective and stable in segmentation tasks. 
As illustrated in Fig.~\ref{fig5}, our decoder aggregates the features from the last three stages, upsamples symmetrically with the encoder, and forms skip connections with multi-scale features, thereby forming an encoder--decoder architecture similar to that of U-Net.


\begin{figure}
\begin{center}
\includegraphics[width=0.6\textwidth]{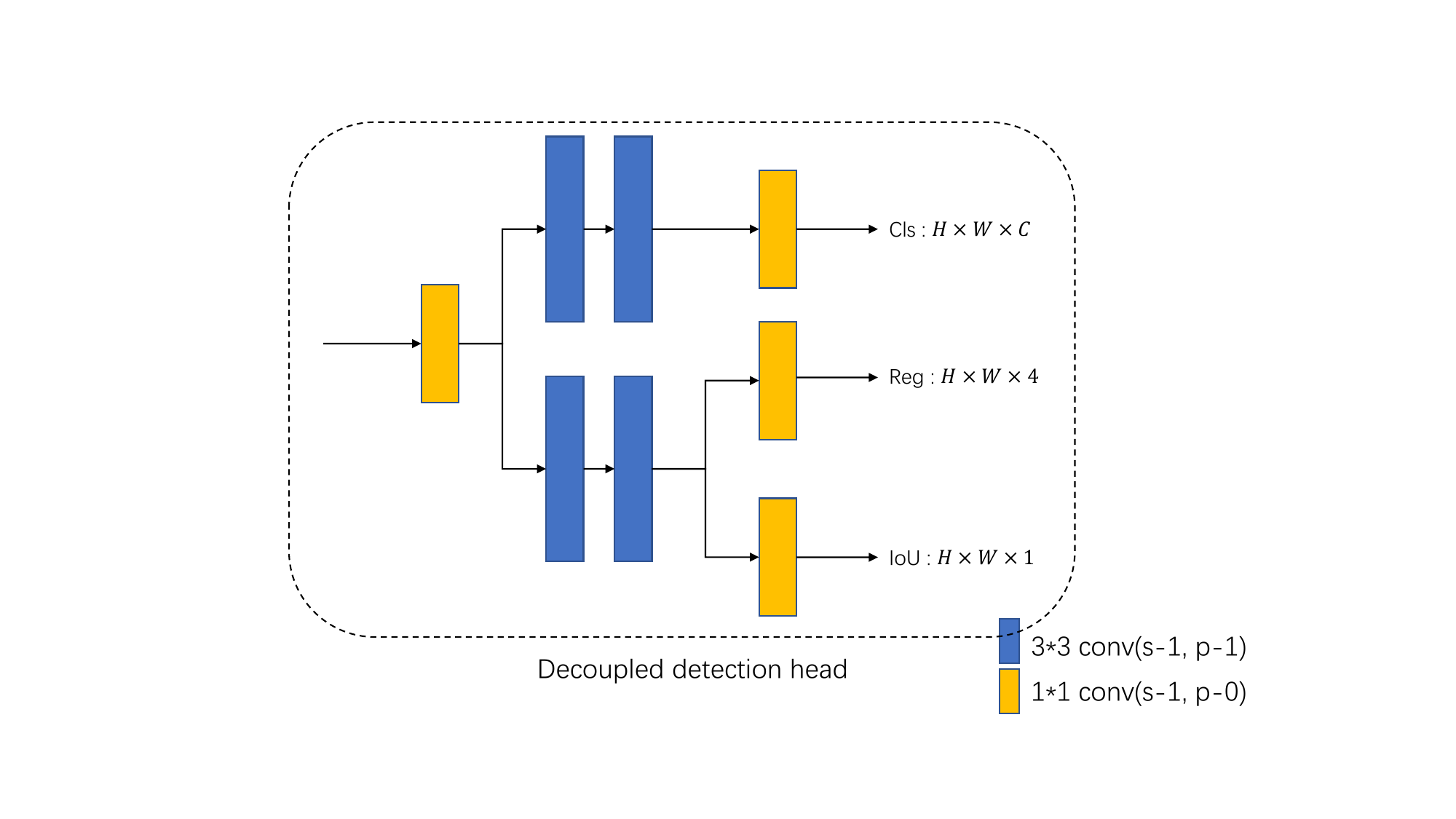}
\end{center}
\caption{Decoupled detection head} \label{fig6}
\end{figure}

\paragraph{Detection Head.} The conflict between the classification and regression tasks is a well-known problem in object detection~\cite{ref_lncs16,ref_lncs17}. In YOLOX~\cite{ref_lncs15}, a lite decoupled detection head separates the classification and regression tasks for end-to-end training, which enables fast fitting for training and improves model performance. As illustrated in Fig.~\ref{fig6}, for each feature stage, this head adopts a 1 × 1 conv layer to reduce the feature channels to 256 and then adds two parallel branches, each with two 3 × 3 conv layers, for the classification and regression tasks, respectively. An IoU branch is added to the regression branch. This decoupled detection head is critical in complex multi-task models.


\subsection{Model Training}

\subsubsection{Public Floor Plan Datasets.}
CVC-FP includes 122 high-resolution images in four different drawing styles, and the image resolution of the dataset ranges from $1098\times 905$ to $7383\times 5671$ pixels~\cite{ref_lncs3}.
R2V~\cite{ref_lncs7} consists of 870 ground-truth floor plan images of urban residences that were collected from various regions in Japan.
R3D~\cite{ref_lncs18} consists of 214 original images, in which most room shapes are irregular with a nonuniform wall thickness.
CubiCasa5k~\cite{ref_lncs8} is a large-scale public dataset that consists of 5000 ground-truth Finnish floor plan images and is annotated using the SVG vector graphics format. 
In this study, we selected CubiCasa5k as the experimental dataset. We think the results of CubiCasa5k is sufficient because of its rich-in-type, high-quality, and close-to-real-world-data nature. 

\subsubsection{Multi-task Loss.} 
Studies~\cite{ref_lncs19} have defined reasonable weights for the loss function from the perspective of the number of labels for segmentation tasks. These loss weights represent the distribution of different target categories in the overall dataset, which can stabilize the loss value during training and enable the model to perform normal iterative training. 

\paragraph{Segmentation-Weighted Loss.} We define the task-weighted loss in an entropy style as follows:

\begin{equation}
L_{segmentation} =  \sum_{i=0}^{C}-\omega _{i}y_{i}logp_{i}
\end{equation}

where $y_i$ is the label of the $i$-th floor plan segmentation element in the floor plan, $C$ is the number of floor plan segmentation elements in the task, 
when $C$ is equal to 0, it represents the background category, in our experiments in this work, when $C$ is equal to 1, it represents the wall category,
and $p_i$ is the prediction label of the pixels for the $i$-th element ($p_i \in [0, 1]$). Furthermore, $w_i$ is defined as follows:

\begin{equation}
w_{i}=\frac{\hat{N}-\hat{N}_{i}}{\sum_{j=0}^{C}(\hat{N}-\hat{N}_{j})}
\end{equation}

where $\hat{N}_{i}$ is the total number of ground-truth pixels for the $i$-th floor plan segmentation element in the floor plan and $\hat{N}=\sum_{i=0}^{C}\hat{N}_{i}$ represents the total number of ground-truth pixels over all $C$ floor plan segmentation elements.
The $W_i$ in the loss function is calculated based on the training-set. Multiple experiments will divide different datasets, the $W_i$ will also be recalculated. The specific value will be different, and the effect of the experiments is stable and consistent. In our experiments in this work, $W_0$ is approximately 0.2, $W_1$ is approximately 0.8.

\paragraph{Detection Loss.}
A widely used loss function combination~\cite{ref_lncs15,ref_lncs20} ensures effective model training in detection tasks. In this study, we use the detection loss functions of YOLOv3~\cite{ref_lncs21}. The overall detection loss consists of the bbox, class, and objectness losses. The segmentation and detection losses are combined as the overall loss during model training to strengthen the relevance of the segmentation and detection task targets.

\begin{equation}
L_{total}=L_{segmentation}+L_{detection}
\end{equation}

\section{Experiments}
We evaluated our method on the public dataset CubiCasa5k. We divided the dataset into 4000 images for training, 500 images for validation, and 500 images for testing. The size of the image data ranged from $430\times 485$ to $6316\times 14304$ with few repeating dimensions, and the average size was $1399\times 1597$. It was necessary for the data to be divisible by 32 to fit the data size change of the model architecture. 
Therefore, we directly resized the input images to $ 1536\times 1536 $, and the aspect ratio will change, but not change greatly. When the input images need to be down-sampled, we use the area interpolation algorithm, and when the input images need to be up-sampled, we use the cubic interpolation algorithm.
The training parameters: the batch size is 10 by default to typical 10-GPU devices, the max lr is 0.01, the initial lr is 0.0001, the min lr is 0.000001, the weight decay is 0.0005 and the SGD momentum is 0.937, and the cosine lr schedule, total epoch is 1000, 50 epochs linear warm-up. After experiments, these parameters can achieve the best training effect in the experimental models. 

CubiCasa5k contains over 80 categories, due to the nature that walls, doors and windows usually jointly constitute the main frame of the architecture of a floorplan, we believe the joint training will have advantages.
In our experiments, the segmentation task was performed on the walls, and the detection task was performed on the doors and windows. All models, including the comparison models, were trained on a node with 10 RTX 3090 GPUs. Because of the particularity of the floor plan data, pretrained models were not used in any of our experiments. That is, all models were trained from scratch. We adopted the AP (@0.5 and @[.5:.95]) and IoU as evaluation metrics for detection and segmentation, respectively. Furthermore, we recorded the convergence speed of the model during the training process.
Regarding the convergence epoch, we define that when the first accuracy of the training epoch reaches 99.9\% of the final, and the subsequent accuracy fluctuations do not exceed 0.2\% of the final, the training curve also indicate the convergence speed.

\subsection{Comparison with Single-Task Models}
U-Net has often been used in segmentation tasks because of its high performance. It is also used for the recognition task of the main architecture in floor plans. Models in the YOLO series are frequently used for fast component detection. We use U-Net and YOLOv3 as the comparative test models to show the performance of MuraNet.

\subsubsection{Comparison with U-Net.} 
\begin{figure}
\begin{center}
\includegraphics[width=0.6\textwidth]{ 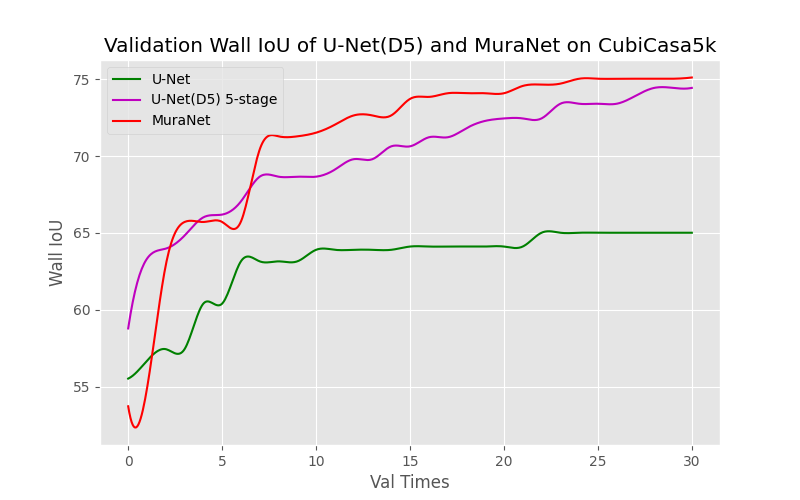}
\end{center}
\caption{ Validation wall IoU of MuraNet and U-Net(D5)} \label{fig1}
\end{figure}

We trained our MuraNet model and U-Net model with base(Fig.~\ref{fig1}), 5, and 6 depth structures using the same training parameter configurations, and the IoU of the wall was considered as the primary evaluation metric.  Since U-Net cannot fit the data well, the data of U-Net 5-stage(D5) which has 5 downsampling modules, one more than the standard U-Net is added in Fig.~\ref{fig1}. As indicated in Table~\ref{tab1}, the wall IoU of MuraNet was generally higher than that of U-Net, which means that MuraNet achieved higher accuracy and the attention-based MURA module can assist the backbone network in improving the segmentation feature extraction. The convergence speed of MuraNet training was faster than that of U-Net at similar performance.

\begin{table}
\centering
\caption{Comparison of MuraNet and U-Net counterparts in terms of IoU (\%) on CubiCasa5k. All models were tested with a resolution of 1536 × 1536
.}\label{tab1}
\begin{tabular}{|l|l|l|}
\hline
Model                &  Wall IoU (\%) & Convergence epochs \\
\hline
U-Net   base         & 65.5          & 6 \\
MuraNet base(+12.9)  & 78.4          & 8 \\
U-Net   5-stage      & 74.4          & 10 \\
MuraNet 5-stage(+1.3)& 75.7          & 9 \\
U-Net   6-stage      & 75.8          & 11 \\
MuraNet 6-stage(+0.6)& 76.4          & 11 \\
\hline
\end{tabular}
\end{table}

\subsubsection{Comparison with YOLOv3.} 
\begin{figure}
\begin{center}
\includegraphics[width=0.6\textwidth]{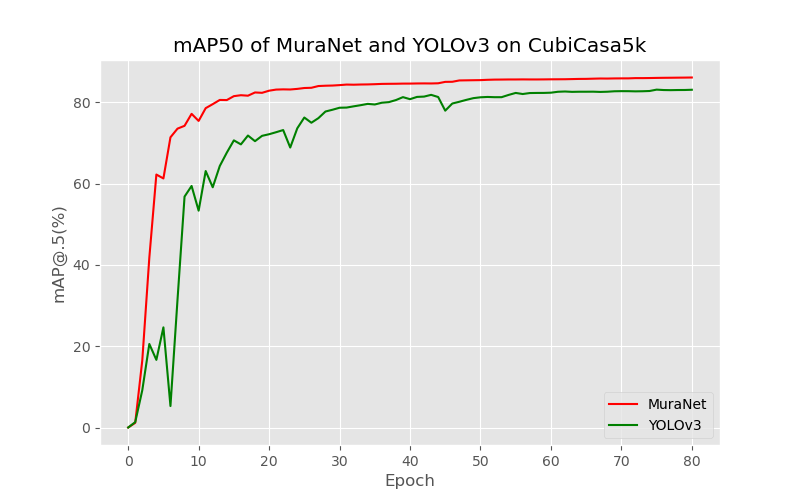}
\end{center}
\caption{ Validation mAP50 of MuraNet and YOLOv3 on CubiCasa5k} \label{fig9}
\end{figure}

We designed MuraNet base(Fig.~\ref{fig9}) and MuraNet-spp, according to the structures of YOLOv3 and YOLOv3-spp, respectively. We trained these models using the same training parameter configuration and used the AP (@0.5 and @0.5:0.95) as the primary evaluation metrics. As indicated in Table~\ref{tab2}, the convergence speed of MuraNet training was generally faster than that of YOLOv3, which means that the decoupled detection head caused the model to converge faster during training. According to the AP value, MuraNet could achieve the same accuracy as YOLOv3 with rapid convergence, and spp-layer has little effect on model performance.

\begin{table}
\centering
\caption{Comparison of MuraNet and YOLOv3 counterparts in terms of AP (\%) on CubiCasa5k. All models were tested with a resolution of 1536 × 1536
.}\label{tab2}
\begin{tabular}{|l|l|l|l|l|l|l|}
\hline

\multirow{2}{*}{Model} & \multicolumn{3}{l|}{ AP50(\%) } & \multicolumn{3}{l|} { AP@[.5:.95](\%) }\\

\cline{2-7}
~       & Doors& Windows & Mean & Doors& Windows & Mean   \\
\hline
YOLOv3 base        & 89.2 & 90.1 & 89.6 & 43.6 & 55.4 & 49.5 \\
MuraNet base(+4.3) & 91.2 & 92.2 & 91.7 & 47.9 & 59.7 & 53.8 \\
YOLOv3+spp         & 89.8 & 90.2 & 90.0 & 43.8 & 55.6 & 49.7 \\
MuraNet+spp(+3.7)  & 90.9 & 91.5 & 91.2 & 47.7 & 59.1 & 53.4 \\
\hline
\end{tabular}
\end{table}

\subsection{Ablation Experiments}

\subsubsection{MURA.} 
\begin{figure}
\begin{center}
\includegraphics[width=0.6\textwidth]{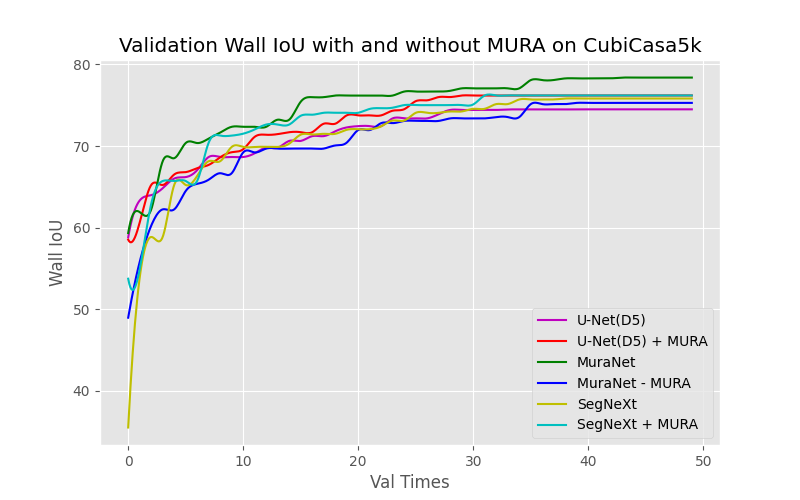}
\end{center}
\caption{ Validation wall IoU of MuraNet, U-Net(D5) and SegNeXt with MURA and without MURA} \label{fig8}
\end{figure}

We designed three comparative experiments to verify the functions of our proposed attention-based MURA module in the backbone for feature extraction from floor plan data. 
In the first experiment, we compared MuraNet model with and without MURA modules.
In the second experiment, we compared the model with U-Net and only added MURA modules to the backbone, without changing the U-Net structure. 
In the third experiment, we compared the model with SegNeXt and replaced the MSCA module, which uses depth-wise strip convolutions to compute the attention, with MURA modules that use a series of $3\times 3$ convolution kernels. 
We also trained these models using the same training parameter configuration on CubiCasa5k. 
As shown in Fig.~\ref{fig8}) and Table~\ref{tab3}, MuraNet, U-Net and SegNeXt with the MURA modules achieved higher wall category accuracy on the CubiCasa5k dataset, and the convergence epochs of the training cycle were approximately the same. Thus, the attention module can extract feature information better, and a series of $3\times 3$ convolution kernels is more suitable for floor plan data than depth-wise strip convolutions.

\begin{table}
\centering
\caption{Comparison of MuraNet, U-Net and SegNeXt with and without MURA modules and counterparts in terms of IoU (\%) on CubiCasa5k. All models were tested with a 1536 × 1536 resolution.}\label{tab3}
\begin{tabular}{|l|l|l|}
\hline
Model                     &  Wall IoU (\%) & Convergence epochs \\
\hline
MuraNet base                & 78.4          & 8  \\
MuraNet-MURA base(-3.1)     & 75.3          & 7 \\
MuraNet 5-stage             & 75.7          & 9 \\
MuraNet-MURA 5-stage(-0.7)  & 75.0          & 10 \\
MuraNet 6-stage             & 76.4          & 11 \\
MuraNet-MURA 6-stage(-0.2)  & 76.2          & 11 \\
\hline
U-Net base                & 65.5          & 6  \\
U-Net+MURA base(+9.58)    & 75.1          & 8 \\
U-Net 5-stage             & 74.4          & 10 \\
U-Net+MURA 5-stage(+1.7)  & 76.1          & 9 \\
U-Net 6-stage             & 75.8          & 11 \\
U-Net+MURA 6-stage(+0.42) & 76.2          & 10 \\
\hline
SegNeXt base              & 75.8          & 12 \\
SegNeXt+MURA base(+0.38)  & 76.2          & 12 \\
SegNeXt small             & 75.3          & 11 \\
SegNeXt+MURA small(+0.71) & 76.0          & 10 \\
SegNeXt large             & 76.3          & 10 \\
SegNeXt+MURA large(-0.02) & 76.3          & 10 \\
\hline
\end{tabular}
\end{table}

\subsubsection{Decoupled Detection Head.} 
\begin{figure}
\begin{center}
\includegraphics[width=0.6\textwidth]{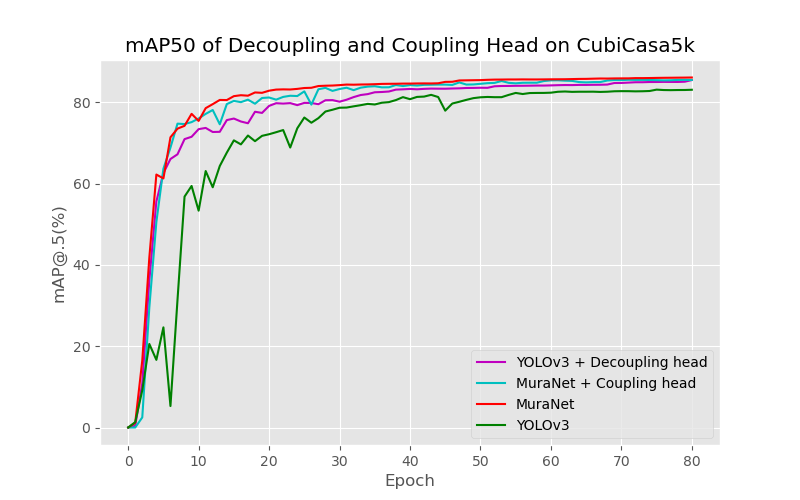}
\end{center}
\caption{ Validation mAP50 of Decoupling and Coupling Head on CubiCasa5k} \label{fig10}
\end{figure}

We only used decoupled and coupled detection heads on MuraNet and YOLOv3 to verify the effect of the decoupled head in the detection task(Fig.~\ref{fig10}). We also trained these models using the same parameter configuration on CubiCasa5k. Table~\ref{tab4} indicates that the decoupled detection head could accelerate the model fitting without compromising on the accuracy in the normal YOLO detection models.

\begin{table}
\centering
\caption{Comparison of MuraNet and YOLOv3 with and without decoupled detection head and counterparts in terms of AP (\%) on CubiCasa5k. All models were tested with a resolution of 1536 × 1536.}\label{tab4}
\begin{tabular}{|l|l|l|l|l|l|l|}
\hline
\multirow{2}{*}{Model} & \multicolumn{3}{l|}{ AP50 (\%) } & \multicolumn{3}{l|}{ AP@[.5:.95] (\%) }\\
\cline{2-7}
~                                & Doors & Windows & Mean  & Doors & Windows & Mean    \\
\hline
MuraNet base+coupled head        & 89.9 & 90.1 & 90.0 & 43.9 & 56.7 & 50.3  \\
MuraNet base+decoupled head(+3.5)& 91.2 & 92.2 & 91.7 & 47.9 & 59.7 & 53.8  \\
MuraNet+spp+coupled head         & 89.7 & 90.1 & 89.9 & 44.0 & 56.2 & 50.1  \\
MuraNet+spp+decoupled head(+3.3) & 90.9 & 91.5 & 91.2 & 47.7 & 59.1 & 53.4  \\
\hline
YOLOv3 base+coupled head        & 89.2 & 90.1 & 89.6 & 43.6 & 55.4 & 49.5  \\
YOLOv3 base+decoupled head(+0.4)& 89.8 & 90.1 & 90.0 & 43.9 & 55.9 & 49.9 \\
YOLOv3+spp+coupled head         & 89.8 & 90.2 & 90.0 & 43.8 & 55.6 & 49.7 \\
YOLOv3+spp+decoupled head(+0.9) & 90.0 & 90.1 & 90.0 & 45.0 & 56.2 & 50.6 \\
\hline
\end{tabular}
\end{table}

\section{Conclusions}
This paper proposes MuraNet, an attention-based multi-task model for segmentation and detection tasks in floor plan data, designed to jointly train the model on both detection and segmentation tasks. MuraMet adopt a unified encoder called MURA module as the backbone, an improved segmentation decoder branch for the segmentation task, and a YOLOX-based decoupled detection head branch for the detection task. The two key contributions of our work is (1) our proposed model integrates the pixel-level segmentation and vector-level detection at the same time. (2) we add attention mechanism for the model to leverage the correlations between walls, doors and windows to enhance the accuracy.

Comparative experiments with U-Net and YOLOv3 on the CubiCasa5k public dataset demonstrate that MuraNet achieves better feature extraction and higher performance by using a unified backbone with the attention mechanism and different head branches for different tasks.
However, we believe that there is room for improvement in MuraNet. While the attention mechanism is currently only utilized in the backbone of MuraNet, we believe that its use in other parts, such as the head or loss function, could lead to better learning of relationships among different recognition targets, improving recognition tasks in floor plan data. This is a direction for future exploration.

%
%
%

\begin{thebibliography}{99}

\bibitem{ref_lncs2}
Dodge, S., Xu, J., Stenger, B.: Parsing floor plan images. In:MVA, pp. 358--361 (2017). \doi{10.23919/MVA.2017.7986875}

\bibitem{ref_lncs3}
de las Heras, L.P., Fernández, D., Valveny, E., Lladós, J., Sánchez, G.: Unsupervised wall detector in architectural floor plans. In: ICDAR, pp. 1245--1249 (2013). \doi{10.1109/ICDAR.2013.252}

\bibitem{ref_lncs4}
Surikov, I.Y., Nakhatovich, M.A., Belyaev, S.Y., et al.: Floor plan recognition and vectorization using combination unet, faster-rcnn, statistical component analysis and Ramer-Douglas-Peucker. In: COMS2, pp. 16--28 (2020)

\bibitem{ref_lncs5}
Wu, Y., Shang, J., Chen, P., Zlantanova, S., Hu, X., and Zhou, Z.: Indoor mapping and modeling by parsing floor plan images. International J. Geogr. Inf. Sci. \textbf{35}(6), 1205--1231 (2021)

\bibitem{ref_lncs6}
Lu, Z., Wang, T., Guo, J., et al.: Data-driven floor plan understanding in rural residential buildings via deep recognition. Inf. Sci.\textbf{567}, 58--74 (2021)

\bibitem{ref_lncs7}
Liu, C., Wu, J., Kohli, P., Furukawa, Y.: Raster-to-vector: revisiting floorplan transformation. In: ICCV, pp. 2195–-2203 (2017)

\bibitem{ref_lncs8}
Kalervo, A., Ylioinas, J., Häikiö, M., Karhu, A., Kannala, J.: CubiCasa5K: A dataset and an improved multi-task model for floorplan image analysis. In: Felsberg, M., Forssén, P.-E., Sintorn, I.-M., Unger, J. (eds.) SCIA 2019. LNCS, vol. 11482, pp. 28–-40. Springer, Cham (2019)

\bibitem{ref_lncs9}
Dosovitskiy, A., et al.: An image is worth 16x16 words: Transformers for image recognition at scale. In: Int. Conf. Learn. Represent (2020)

\bibitem{ref_lncs10}
Guo, M.H., Lu, C.Z., Liu, Z.N., Cheng, M.M., Hu, S.M.: Visual Attention Network. arXiv preprint arXiv:2202.09741 (2022)

\bibitem{ref_lncs11}
Guo, M.H., et al.: SegNeXt: Rethinking convolutional attention design for semantic segmentation. arXiv preprint arXiv:2209.08575 (2022)

\bibitem{ref_lncs12}
Ronneberger, O., Fischer, P., Brox, T.: U-Net: convolutional networks for biomedical image segmentation. In: MICCAI (2015)

\bibitem{ref_lncs13}
Xie, E., Wang, W., Yu, Z., Anandkumar, A., Alvarez, J.M., Luo, P.: Segformer: Simple and efficient design for semantic segmentation with transformers. Adv. Neural Inform. Process. Syst.34 (2021)

\bibitem{ref_lncs14}
Chen, L.C., Papandreou, G., Kokkinos, I., Murphy, K., Yuille, A.L.: Deeplab: Semantic image segmentation with deep convolutional nets, atrous convolution, and fully connected crfs. IEEE Trans. Pattern Anal. Mach. Intell. 40(4), 834--848 (2018)

\bibitem{ref_lncs15}
Ge, Z., Liu, S., Wang, F., Zeming, L., Jian, S.: YOLOX: Exceeding YOLO Series in 2021. arXiv preprint arXiv:2107.08430 (2021)

\bibitem{ref_lncs16}
Song, G., Liu, Y., Wang, X.: Revisiting the sibling head in object detector. In: CVPR (2020)

\bibitem{ref_lncs17}
Wu, Y, Chen, Y., Yuan, L. et al.: Rethinking classification and localization for object detection: In: CVPR (2020)

\bibitem{ref_lncs18}
Liu, C., Schwing, A., Kundu, K., Urtasun, R., and Fidler, S.: Rent3D: Floor-plan priors for monocular layout estimation. In: CVPR (2015)

\bibitem{ref_lncs19}
Zeng, Z., Li, X., Yu, Y.K., Fu, C.W.: Deep floor plan recognition using a multi-task network with room-boundary-guided attention. In: ICCV, pp. 9095--9103 (2019)

\bibitem{ref_lncs20}
Ge, Z., Liu, S., Li, Z., Yoshie, O., and  Sun, J.: Ota: Optimal transport assignment for object detection. In CVPR, pp. 303--312 (2021)

\bibitem{ref_lncs21}
Redmon, J., Farhadi, A.: YOLOv3: An incremental improvement. arXiv preprint arXiv:1804.02767 (2018)

\bibitem{ref_lncs22}
Zhao, Y., Xueyuan, D., and Huahui, L.: A deep learning-based method to detect components from scanned structural drawings for reconstructing 3D models. Appl. Sci. 10.6: 2066 (2020)

\bibitem{ref_lncs23}
Rezvanifar, A., Cote, M., and Albu, A.B.: Symbol spotting on digital architectural floor plans using a deep learning-based framework. In: CVPRW (2020)

\bibitem{ref_lncs24}
Fan, Z., Zhu, L., Li, H., et al.: FloorPlanCAD: a large-scale CAD drawing dataset for panoptic symbol spotting. In: ICCV (2021)

\bibitem{ref_lncs25}
Nicolas, C., Francisco, M., Gabriel, S., Nicolas, U., Alexander, K., Sergey, Z.:  End-to-End Object Detection with Transformers. arXiv:2005.12872 (2020)

\bibitem{ref_lncs26}
Ze, L., Yutong, L., Yue, C., et al.: End-to-End Object Detection with Transformers. In: ICCV (2021)

\bibitem{ref_lncs27}
Wang, J., Sun, K., Cheng, T., et al.: Deep high-resolution representation learning for visual recognition. IEEE transactions on pattern analysis and machine intelligence 43.10: 3349--3364 (2020) 

\bibitem{ref_lncs28}
Liang-Chieh, C., George, P., Iasonas, K., Kevin, M., Alan, L.Y.: Semantic image segmentation with deep convolutional nets and fully connected crfs. arXiv preprint arXiv:1412.7062 (2014) 

\bibitem{ref_lncs29}
He, K., Gkioxari, G., Dollár, P., Girshick, R.: Mask r-cnn. In: ICCV(2017) 

\bibitem{ref_lncs30}
Ren, S., He, K., Girshick, R., Sun, J.: Faster R-CNN: Towards Real-Time Object Detection with Region Proposal Networks IEEE Trans. Pattern Anal. Mach. Intell. 39, 1137–-1149 (2017)


\end{thebibliography}
%

\end{document}